\documentclass[journal]{IEEEtran}

\usepackage{times}
\usepackage{epsfig}
\usepackage{graphicx}
\usepackage{subfigure}
\usepackage{amsmath}
\usepackage{amssymb}
\usepackage[percent]{overpic}
\usepackage{cite}
\usepackage{verbatim}
\usepackage{bm}

\usepackage[pagebackref=true,breaklinks=true,letterpaper=true,colorlinks,bookmarks=false]{hyperref}
\usepackage{cleveref}
\usepackage{multirow}
\usepackage{stfloats}
\usepackage{marvosym}

\usepackage{algorithmic}

\usepackage{makecell}

\usepackage[linesnumbered, ruled]{algorithm2e}
\SetKwRepeat{Do}{do}{while}
\usepackage{colortbl}
\usepackage{pifont}

\usepackage{booktabs}

\ifCLASSINFOpdf

\else

\fi

\begin{document}

\title{Motion-Aware Vision-Reference Alignment for Referring Multi-Object Tracking}

\author{Weiyi~Lv,
        Ning~Zhang,
        Hanyang~Sun,
        Haoran~Jiang,
        Kai~Zhao,
        Yixiao~~Gu,
        Jing~Xiao,~\IEEEmembership{Fellow,~IEEE,}
        and~Dan~Zeng\textsuperscript{\Letter},~\IEEEmembership{Senior~Member,~IEEE}%

\thanks{This work was supported in part by the National Natural Science Foundation of China under Grant 62372284 and Shanghai Natural Science Foundation under Grant 24ZR1421800.}
\thanks{\Letter Dan Zeng is the corresponding author.}
\thanks{W. Lv, H. Sun, H. Jiang, Y. Gu, D. Zeng, and K. Zhao are with the School of Communication and Information Engineering, Shanghai University, Shanghai 200444, China (e-mail: \{kroery, sunhanyang, jianghaoran, yxgu, dzeng\}@shu.edu.cn, kz@kaizhao.net).}
\thanks{N. Zhang and J. Xiao are with PAII Inc., Palo Alto, CA, 94306, USA. (e-mail: ning.zhang@gmail.com, xiaojing661@pingan.com.cn).}
\thanks{%
  \copyright~2026 IEEE. Personal use of this material is permitted.
  Permission from IEEE must be obtained for all other uses, in any
  current or future media, including reprinting/republishing this
  material for advertising or promotional purposes, creating new
  collective works, for resale or redistribution to servers or lists,
  or reuse of any copyrighted component of this work in other works.}
}

\maketitle

\begin{abstract}
Referring Multi-Object Tracking (RMOT) extends conventional multi-object tracking (MOT) by introducing natural language references for multi-modal fusion tracking.
RMOT benchmarks only describe the object’s appearance, relative positions, and initial motion states. This so-called static regulation fails to capture dynamic changes of the object motion, including velocity changes and motion direction shifts. This limitation not only causes a temporal discrepancy between static references and dynamic vision modality but also constrains multi-modal tracking performance.
To address this limitation, we propose a novel motion-aware vision-reference alignment framework, named VMRMOT. 
VMRMOT introduces a motion modality derived from object dynamics to facilitate the alignment between the vision modality and language references.
Specifically, we introduce motion-aware descriptions derived from object dynamic behaviors and encode them into motion features as the motion modality through multi-modal large language models (MLLMs).
We further design a Vision-Motion-Reference Alignment (VMRA) module to hierarchically align visual queries with motion and reference cues, enhancing their cross-modal consistency.
In addition, a Motion-Guided Prediction Head (MGPH) is developed to explore motion modality to enhance the performance of the prediction head.
To the best of our knowledge, VMRMOT is the first motion-aware vision-reference alignment framework for the RMOT task.
Extensive experiments on multiple RMOT benchmarks demonstrate that VMRMOT outperforms existing state-of-the-art methods. The code is available at \href{https://github.com/Kroery/VMRMOT}{https://github.com/Kroery/VMRMOT}.

\end{abstract}


\IEEEpeerreviewmaketitle


\begin{figure*}[t]
\centering
\begin{overpic}
[width=1\linewidth,]{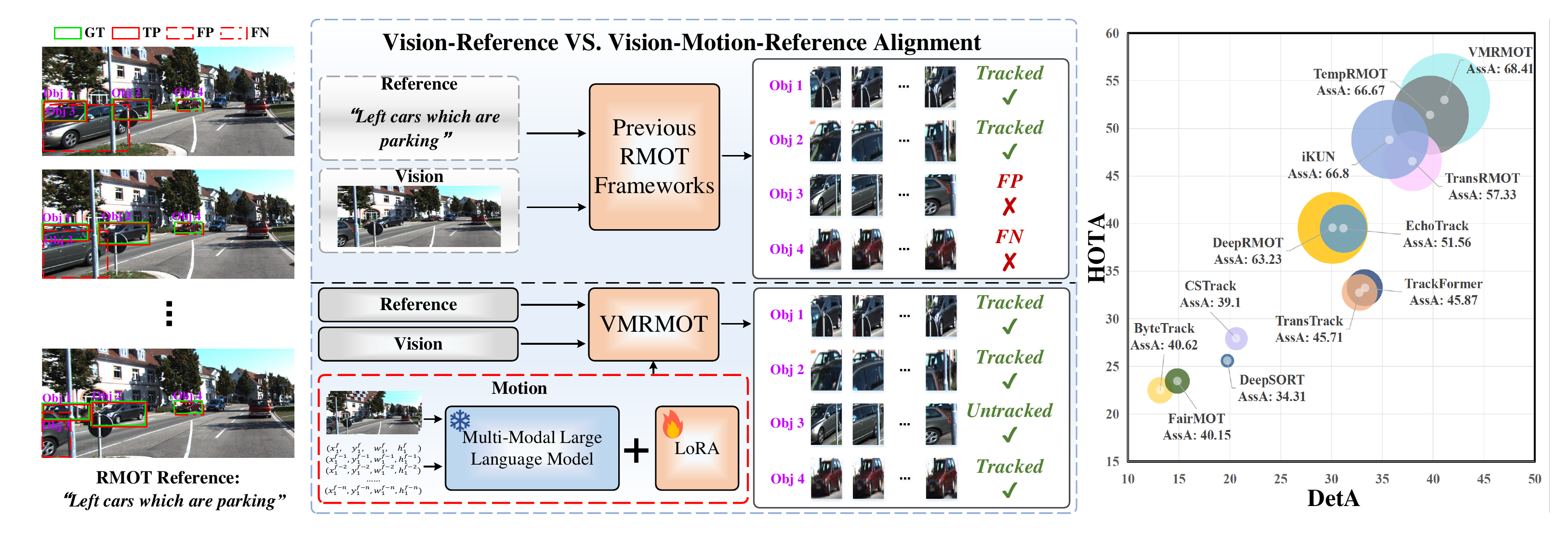}
\put(2,-1){\footnotesize{(a)} Temporal discrepancy}
\put(31,-1){\footnotesize{(b)} Comparison of different pipelines}
\put(76,-1){\footnotesize{(c)} Performance comparison}

\end{overpic}
\caption{
(a) illustrates the temporal discrepancy between the static reference “left cars which are parking” and the dynamic vision modality of objects 3 and 4, which leads to FP and FN.
(b) shows the comparison between previous RMOT frameworks, which integrate only the vision and reference modalities, and our proposed VMRMOT, which incorporates vision, motion, and reference modalities. 
(c) shows the HOTA-DetA-AssA comparisons of different RMOT trackers on the Refer-KITTI dataset. 
Our VMRMOT achieves $53.00\%$ HOTA, $41.13\%$ DetA, and $68.41\%$ AssA.
}
\vspace{-10pt}
\label{fig1}
\end{figure*}

\section{Introduction}
\label{sec1}
Referring Multi-Object Tracking (RMOT) is a recent active research branch of conventional multi-object tracking (MOT) \cite{(3)bytetrack, (35)pidmot, (36)diffmot, (37)motip, (39)memotr}, which introduces natural language references as an additional modality to specify the objects of interest (e.g., “\textit{Track left vehicles which are parking}”).
Unlike traditional MOT that relies only on visual appearance and motion cues, RMOT enables human-centered and instruction-driven tracking, bridging perception with natural communication. 
Benefiting from the integration of vision and reference modalities, RMOT has attracted increasing attention in various applications such as video editing \cite{(32)editing}, autonomous driving \cite{(33)driving, (47)driving2}, and animal surveys \cite{(34)animal}.

Existing RMOT methods \cite{(10)transrmot, (7)ikun, (11)temprmot} encode the features of language references and fuse them directly with visual representations, either at early feature extraction stages or using them later to select tracking results.
However, current RMOT benchmarks formulate language references through static regulations, which are manually defined rules or templates that describe objects without considering their temporal dynamics. Such static references typically encode static attributes (e.g., color, category, or relative position) or initial motion states (e.g., initial velocity or starting location). 
On the other hand, the vision modality encompasses the evolving behaviors of objects over time, including changes in velocity, (de/ac)celeration, and abrupt motion direction changes.
Since static references cannot fully represent such temporal dynamics, existing methods fail to achieve accurate cross-modal alignment.
This is due to their exclusive dependency on static references.
As illustrated in Fig.~\ref{fig1} (a), the temporal discrepancy between static references and the vision modality often results in false positives (FP) and false negatives (FN), thereby limiting the effectiveness of multi-modal tracking.

Recognizing the importance of aligning static references with the vision modality, some efforts \cite{(24)dkgtrack, (26)CDRMT, (27)sktrack} have begun to explore the decoupling of language references into finer-grained components.
Despite these advances, such methods still rely on static references.
Consequently, they do not account for the dynamic motion variations
in object behavior observed across video sequences, limiting their ability to achieve precise vision–reference temporal alignment.

In this paper, we propose a novel motion-aware vision-reference alignment RMOT framework named VMRMOT. 
It aims to bridge the discrepancy between static language references and the vision modality.
As shown in Fig.~\ref{fig1} (b), different from previous RMOT frameworks, VMRMOT introduces a motion modality to facilitate vision–reference alignment.
Specifically, we first generate motion-aware descriptions that explicitly encode the dynamic behaviors of objects. 
From each object’s historical trajectory, we extract key motion cues including position, motion direction, distance trend, and speed trend, and organize them into compact natural-language descriptions. These motion descriptions effectively compress historical motion into a compact textual representation, capturing temporal dynamics that static references cannot.
Then, we employ MLLMs to encode the motion descriptions into motion features as the motion modality.
To hierarchically fuse motion and reference modalities into the vision modality, we design a Vision–Motion–Reference Alignment (VMRA) module that sequentially integrates motion and reference cues into visual queries.
This hierarchical fusion allows the model to first incorporate temporal motion cues, then refine alignment with static language semantics, ensuring robust multi-modal tracking.
Furthermore, we introduce a Motion-Guided Prediction Head (MGPH), which leverages motion modality to enhance the performance of the prediction head.

To the best of our knowledge, VMRMOT is the first motion-aware vision–reference alignment framework for RMOT.
Extensive experiments on multiple RMOT benchmarks demonstrate that our method consistently outperforms state-of-the-art approaches. As we can see in Fig.~\ref{fig1} (c), VMRMOT achieves the best overall performance in terms of HOTA, DetA, and AssA metrics on the Refer-KITTI dataset.
The results confirm that introducing the motion modality effectively enhances the alignment between reference and vision modalities, leading to improved cross-modal tracking performance.

In summary, our contribution is three-fold:

\begin{itemize}

\item We propose a novel RMOT framework, VMRMOT, which introduces a motion modality to enhance vision–reference alignment.

\item We design a VMRA module and an MGPH. The VMRA leverages motion cues to enhance the alignment between vision and reference features,  while the MGPH leverages motion modality to enhance the performance of the prediction head.

\item VMRMOT outperforms SOTA methods on multiple public RMOT benchmarks, consistently improving HOTA, DetA, and AssA metrics.

\end{itemize}

\begin{figure*}[t]
\centering
\begin{overpic}
[width=1\linewidth,]{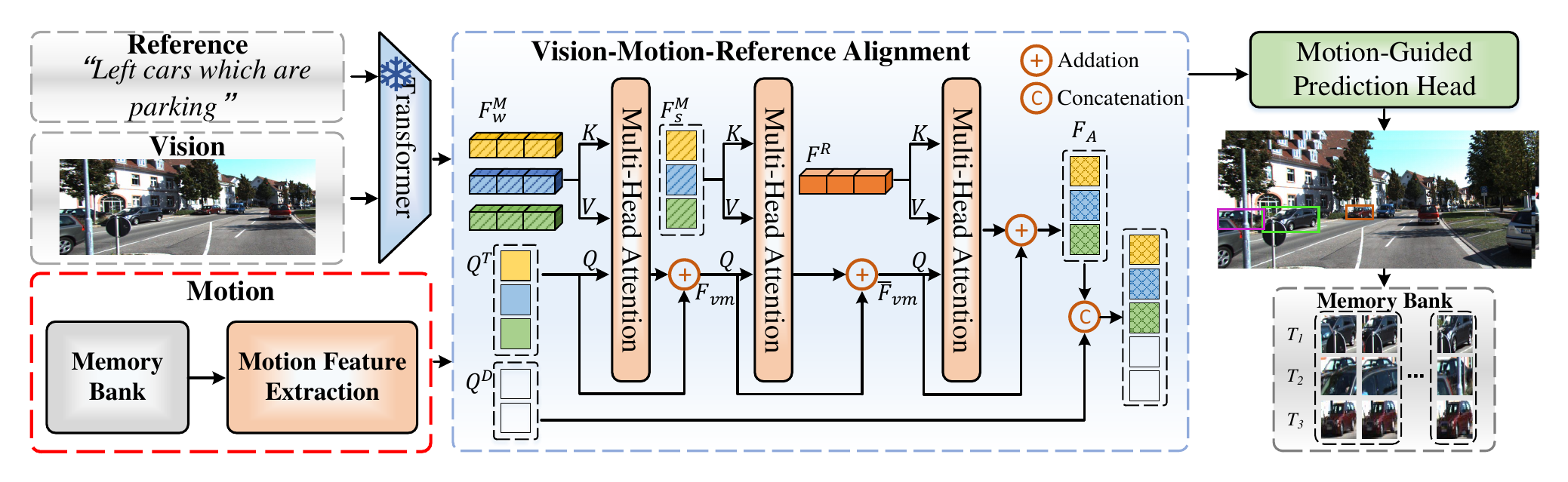}

\end{overpic}
\vspace{-15pt}
\caption{The overall architecture of VMRMOT. VMRMOT consists of four parts: a frozen transformer, a motion feature extraction, a vision–motion–reference alignment, and a motion-guided prediction head.
}
\vspace{-10pt}
\label{fig2}
\end{figure*}

\section{Related Work}
\label{sec2}

\subsection{Referring Multi-Object Tracking}
Recently, referring understanding \cite{(19)dkg6, (20)dkg16, (21)dkg42, (22)dkg47, (23)dkg50, (44)referdino, (40)Refer-Youtube-VOS, (41)trackgpt25, (42)trackgpt27, (43)trackgpt51} has attracted great attention, benefiting from its flexibility and alignment with human intentions.
As an emerging referring understanding task, Referring Multi-Object Tracking (RMOT) requires models to localize and consistently track objects specified by natural language references across video frames.

Early RMOT approaches are largely built upon end-to-end multi-object tracking frameworks. 
TransRMOT \cite{(10)transrmot} introduces RMOT by extending MOTR \cite{(17)motr} with an early visual–reference fusion module. 
TempRMOT \cite{(11)temprmot} integrates temporal information into the model on top of TransRMOT to improve tracking stability. 
In addition to end-to-end frameworks, several studies adopt a two-stage paradigm, where object tracklets are first generated using an off-the-shelf tracker and then matched with language references. 
For instance, iKUN \cite{(7)ikun} first extracts object tracklets and subsequently performs reference matching to identify object trajectories. 
Other works \cite{(9)deeprmot, (25)mlstrack, (28)tenrmot, (38)CGATracker, (48)mglt} follow similar paradigms, either injecting reference embeddings into visual features at early stages or utilizing reference cues to filter tracking outputs at later stages. 

Despite promising progress, most existing methods treat language references as holistic embeddings and primarily rely on feature fusion strategies. While such designs improve visual–language interaction, they often overlook the explicit modeling of dynamic motion cues that are crucial in dynamic tracking scenarios.

\subsection{Motion Modeling in RMOT}

Motion modeling plays a critical role in tracking tasks, as temporal consistency and dynamic behaviors provide essential cues for object association. In MOT, motion cues \cite{(51)motiontrack, (52)deft, (53)flow, (36)diffmot} are often explicitly modeled to guide tracking decisions. For instance, DiffMOT \cite{(36)diffmot} directly leverages consecutive object trajectories as motion features to guide diffusion-based generation for tracking. 
MotionTrack \cite{(51)motiontrack} encodes continuous spatial coordinates in image space to model object dynamics across frames. Such approaches treat motion as a dedicated representation derived from trajectory evolution, demonstrating its effectiveness in handling occlusions, abrupt movements, and complex interactions.

In RMOT, however, motion information is typically captured implicitly through temporal aggregation modules, rather than being explicitly formulated as an independent modality. 
Recent works recognize that motion-related descriptions can play distinct roles at different tracking stages. Motivated by this insight, some efforts \cite{(24)dkgtrack, (26)CDRMT, (27)sktrack} begin to explore the decoupling of language references, separating motion-relevant cues from other semantic components.
DKGTrack \cite{(24)dkgtrack} decouples language references into localized descriptions and motion states for precise object tracking. 
CDRMT \cite{(26)CDRMT} emulates the "what" and "where" pathways from the human visual processing system to cognitively decouple references.
However, these methods still rely on decoupling within references. Such references, formulated through static regulations, primarily describe the objects’ static attributes or initial motion states, making them difficult to align with the dynamic vision modality.
In this paper, we explicitly extract motion modality from the dynamic behaviors of objects to facilitate alignment between the vision modality and static references.
Unlike traditional motion priors that directly operate on geometric trajectories for association or prediction, our motion modality reformulates object dynamics into structured motion-aware descriptions and utilizes them for vision--reference alignment. This design enables motion information to serve as an explicit bridge between static references and dynamic vision modality.

\subsection{MLLMs in Video Referring Understanding}

Multimodal Large Language Models (MLLMs) have recently demonstrated strong cross-modal reasoning and representation capabilities, enabling significant progress in video referring and segmentation tasks \cite{(29)trackgpt, (30)visa, (31)refergpt, (45)villa, (46)mevis}. 

TrackGPT \cite{(29)trackgpt} pioneers the integration of MLLMs for reasoning-driven video object segmentation. 
VISA \cite{(30)visa} leverages MLLMs to generate temporally coherent mask sequences conditioned on implicit textual queries, incorporating world knowledge and contextual reasoning. 
ReferGPT \cite{(31)refergpt} proposes a zero-shot RMOT framework that utilizes MLLMs to generate object descriptions for reference matching.

These methods primarily employ MLLMs for reasoning, description generation, or zero-shot inference. In contrast, our approach leverages MLLMs as motion-aware encoders to model motion descriptions derived from visual dynamics. Rather than using MLLMs solely for high-level reasoning, we utilize their cross-modal representational capacity to encode motion as an explicit modality, enabling alignment among motion, vision, and references.

\section{Method}
\label{sec3}

\subsection{Framework Overview}
\label{sec:3.1}
In this section, we introduce VMRMOT, a novel framework that achieves vision–motion–reference alignment for RMOT.
As illustrated in Fig.~\ref{fig2}, VMRMOT follows the end-to-end transformer-based tracking architecture \cite{(17)motr}, and consists of four main components: a frozen transformer, a motion feature extraction, a vision–motion–reference alignment, and a motion-guided prediction head.

\textbf{Transformer.}
Given a video sequence $V$
and a natural language reference $\mathcal{E}^{R}$.
VMRMOT adopts a transformer-based tracker following the design of TransRMOT \cite{(10)transrmot}.  
A ResNet-50 backbone is employed to extract the visual features from each input frame, 
while a pre-trained RoBERTa model \cite{(18)roberta} encodes the references to obtain reference embeddings $F^{R}\in\mathbb{R}^{L^R\times D}$, where $L^R$ refers to the length of reference words and $D$ refers to the feature dimension.   
An early-fusion module integrates the reference and visual features, allowing the tracker to inject reference semantics into the visual features at an early stage.  
The fused features are then fed into a stack of deformable transformer encoder and decoder layers, 
which perform cross-modal interaction and temporal reasoning to associate objects across frames. 
Finally, the transformer outputs a set of detection queries $Q^{D}$ and track queries $Q^{T}$, which incorporate preliminary reference information.  
These queries serve as the initial representations for object localization and identity association.

\textbf{Motion Feature Extraction.}
For each track query, the historical trajectories stored in a memory bank $\mathcal{M}$ are first retrieved and converted into motion-aware descriptions $\mathcal{E}^{M}$.  
These descriptions summarize the dynamic behaviors of objects, including position, motion direction, distance trends, and speed trends.  
The motion-aware descriptions and the previous $n$ frames are jointly fed into MLLMs to obtain motion features as the motion modality, which capture dynamic motion patterns and their temporal consistency with visual sequences.

\textbf{Vision–Motion–Reference Alignment.}
The motion features are integrated with the track queries and reference features within the VMRA module. 
The track queries hierarchically align with motion and reference cues and yield motion- and reference-enhanced visual representations.

\textbf{Motion-Guided Prediction Head.}
The MGPH leverages the motion modality to refine 
both the box and the referring branches,
enhancing the accuracy and consistency of the predicted trajectories. 
The predicted object states are then updated in the memory bank $\mathcal{M}$ to provide historical motion information for subsequent tracking.

The detailed motion feature extraction process is described in Sec.~\ref{sec:3.2}, and the designs of VMRA and MGPH are presented in Sec.~\ref {sec:3.3} and Sec.~\ref{sec:3.4}, respectively.

\begin{figure}[t]
\centering
\begin{overpic}
[width=1\linewidth,]{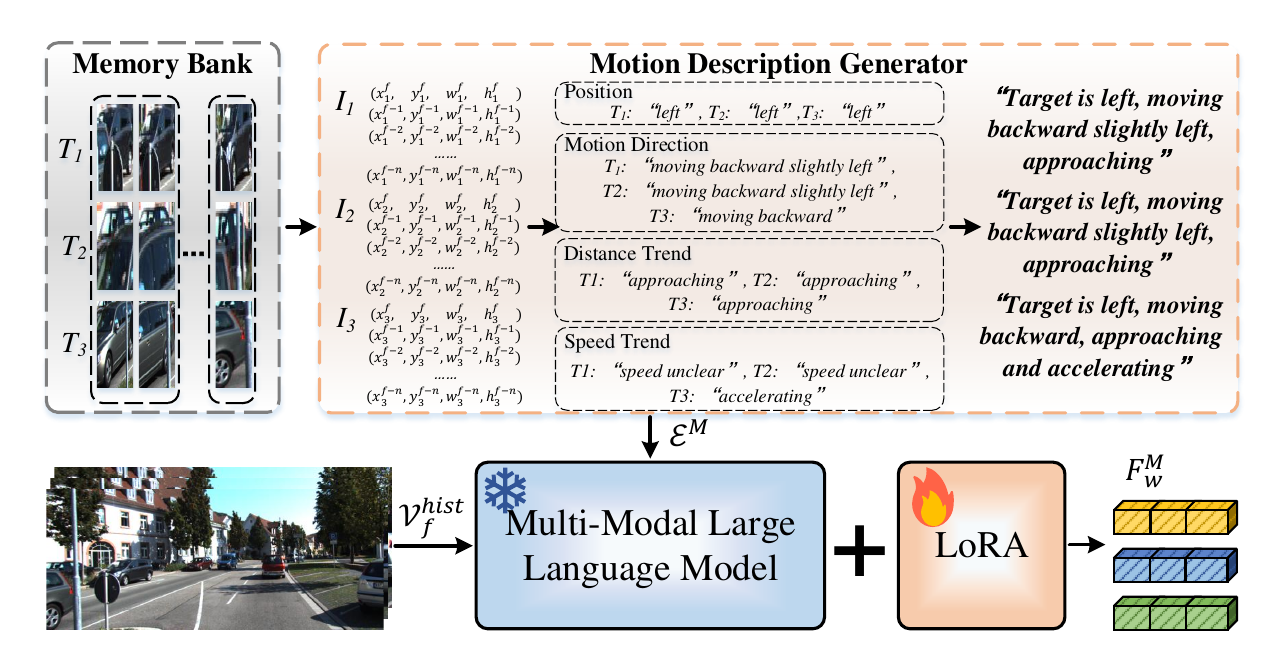}

\end{overpic}
\vspace{-5pt}
\caption{Illustration of the Motion feature extraction pipeline. It consists of two stages: first, historical trajectories are converted into compact motion-aware descriptions; then, MLLMs are employed to extract motion features from these descriptions.}
\vspace{-10pt}
\label{fig3}
\end{figure}

\subsection{Motion Feature Extraction via MLLMs}
\label{sec:3.2}
The purpose of the Motion feature extraction is to obtain motion features from the historical trajectories of tracked objects, which are utilized as the motion modality in the proposed framework.
The pipeline is illustrated in Fig.~\ref
{fig3}.
First, historical trajectories are converted into compact motion-aware descriptions that capture the semantic evolution of object movements. 
Then, leveraging the cross-modal representation capabilities of MLLMs, the motion descriptions are transformed into motion features that encode dynamic motion information and provide a structured embedding of object dynamics.

\textbf{Motion-aware Description Generation.}
To comprehensively characterize the dynamic behaviors of tracked objects, we describe their temporal evolution from four complementary aspects: position, motion direction, distance trend, and speed trend.
These aspects jointly capture both spatial and temporal variations — where the position reflects the object’s relative location, the motion direction indicates its movement orientation, the distance trend reveals its approach or departure from the camera, and the speed trend describes the acceleration or deceleration patterns.

For each tracked object $T_i$, we first retrieve its historical trajectory from the memory bank. 
Let $I=\{B_{f-n}, B_{f-n+1}, \cdots, B_{f}\}$ denote the historical motion information of a tracked object, where $f$ is the frame index, and $B_{f}=(x_{f}, y_{f}, w_{f}, h_{f})$ is the object's bounding box representing the coordinates of the center point and the height and width of the box. 
The historical motion information $I$ is thus defined as the sequence of bounding boxes from the previous $n$ frames, which encapsulates the object’s recent spatial and scale variations.

Based on $I$, we compute four motion descriptors to describe the object’s dynamic behaviors. 
The \textbf{position} descriptor is determined by the normalized center coordinates of the bounding box:
\begin{equation}
P_f = \left( \frac{x_f}{W}, \frac{y_f}{H} \right),
\label{eq1}
\end{equation}
where $W$ and $H$ denote the frame width and height, respectively. 
The position descriptor is then converted into a textual phrase $<$\textit{position}$>$, such as ``\textit{ahead}'', ``\textit{left}'', or ``\textit{right}''.
The \textbf{motion direction} descriptor is computed from the displacement between consecutive frames:
\begin{equation}
\Delta x_f = x_f - x_{f-1}, \quad \Delta y_f = y_f - y_{f-1},
\label{eq2}
\end{equation}
and is converted into a textual phrase $<$\textit{motion direction}$>$, such as ``\textit{moving forward}'', or ``\textit{moving backward}''.
The \textbf{distance trend} descriptor captures changes in the object’s scale:
\begin{equation}
A_f = w_f \cdot h_f, \quad \Delta A_f = A_f - A_{f-1},
\label{eq3}
\end{equation}
The distance trend is then expressed as a textual phrase $<$\textit{distance trend}$>$, such as ``\textit{approaching}'', ``\textit{moving away}'', or ``\textit{staying at distance}''.
The \textbf{speed trend} descriptor is derived from the change in speed magnitude:
\begin{equation}
v_f = \sqrt{(\Delta x_f)^2 + (\Delta y_f)^2}, \quad \Delta v_f = v_f - v_{f-1},
\label{eq4}
\end{equation}
and is converted into a textual phrase $<$\textit{speed trend}$>$, such as ``\textit{accelerating}'', ``\textit{decelerating}'', or ``\textit{constant speed}''.

Finally, the four descriptors are combined to form the motion-aware description.  
The motion description $\mathcal{E}^{M}$ follows the template: 
\begin{center}
\small
\textit{Target is $<$position$>$, $<$motion direction$>$, $<$distance trend$>$ [and $<$speed trend$>$].}
\end{center}

\textbf{Motion Feature Extraction.} 
To obtain motion features as the motion modality, we leverage MLLMs to transform motion descriptions into representations that encode dynamic motion information.
Specifically, for each tracked object at frame $f$, the motion-aware description $\mathcal{E}^{M}$ consisting of $L^M$ words and the visual sequence consisting of the previous $n$ frames
$\mathcal{V}^{hist}_f = \{ \mathbf{F}^{vis}_{f-n}, \mathbf{F}^{vis}_{f-n+1}, \dots, \mathbf{F}^{vis}_f \}$
where $ \mathbf{F}^{vis}_i$ denotes the image of frame $i$, are fed into a multi-modal large language model $\operatorname{MLLM}(\cdot)$:
\begin{equation}
F^{M}_w = \operatorname{MLLM}(\mathcal{E}^{M}_f, \mathcal{V}^{hist}_f),
\label{eq5}
\end{equation}
where $F^{M}_w\in\mathbb{R}^{L^M\times D}$ represents the word-level motion features output by the last layer of the MLLM, which can be further average-pooled to obtain the sentence-level motion feature $F^{M}_s\in\mathbb{R}^{1\times D}$.
The MLLM is jointly trained with the overall VMRMOT framework and LoRA-finetuned specifically on the text encoding and multi-modal fusion modules to better process motion-aware descriptions, which differ from conventional natural language references.

\subsection{Vision-Motion-Reference Alignment}
\label{sec:3.3}
As shown in Fig.~\ref{fig2}, 
we design a VMRA module that hierarchically aligns visual queries with motion and reference embeddings to enhance cross-modal consistency.
Given the visual track query $Q^{T}$, the sentence-level motion feature $\mathbf{F}^{M}_s$, the word-level motion feature $\mathbf{F}^{M}_w$, and the
reference feature $\mathbf{F}^{R}$, the VMRA module hierarchically fuses these features through multi-stage cross-attention to produce a unified embedding.
We perform the VMRA exclusively on track queries, and detection queries are separated and remain unaffected during this process.

The VMRA first modulates the interaction between the visual query $Q^{T}$ and the fine-grained motion semantics $\mathbf{F}^{M}_w$, allowing the model to perceive detailed temporal variations such as "\textit{moving left}", "\textit{accelerating}", or "\textit{approaching}".
This process is formulated as:
\begin{equation}
\mathbf{F}_{vm} = \operatorname{MHA}(Q^{T},\mathbf{F}^{M}_w,\mathbf{F}^{M}_w) + Q^{T},
\label{eq6}
\end{equation}
where $\operatorname{MHA}(\cdot)$ denotes multi-head attention.
Through this step, the visual query dynamically aggregates fine-grained motion cues, enhancing temporal awareness.

Next, 
the vision-motion feature $\mathbf{F}_{vm}$ interacts with the sentence-level motion embeddings $\mathbf{F}^{M}_s$ through another cross-attention layer:
\begin{equation}
\overline{\mathbf{F}}_{vm} = \operatorname{MHA}(\mathbf{F}_{vm}, \mathbf{F}^{M}_s, \mathbf{F}^{M}_s) + \mathbf{F}_{vm},
\label{eq7}
\end{equation}
where the attention is computed across all tracked objects, allowing each query position to access the sentence-level motion summaries of other objects.
This design enables global motion interactions that enhance consistency among correlated trajectories, 
thereby strengthening the coherence between local dynamics and overall motion semantics.

Finally, the vision-motion-enhanced feature $\overline{\mathbf{F}}_{vm}$ is aligned with the reference features $\mathbf{F}^{R}$ to ensure semantic consistency across modalities:
\begin{equation}
\mathbf{F}_{A} = \operatorname{MHA}(\overline{\mathbf{F}}_{vm}, \mathbf{F}^{R}, \mathbf{F}^{R}) + \overline{\mathbf{F}}_{vm}.
\label{eq8}
\end{equation}
The resulting multi-modal feature $\mathbf{F}_A$ is then used to update the original track query $Q^{T}$, producing a motion- and reference-aware query that integrates vision, motion, and reference modalities. 
The updated track queries are then concatenated with the detection queries and fed into the MGPH. 
Through this hierarchical fusion, the VMRA module produces temporally coherent and semantically enriched features that serve as a robust foundation for the subsequent prediction head.

\begin{figure}[t]
\centering
\begin{overpic}
[width=1\linewidth,]{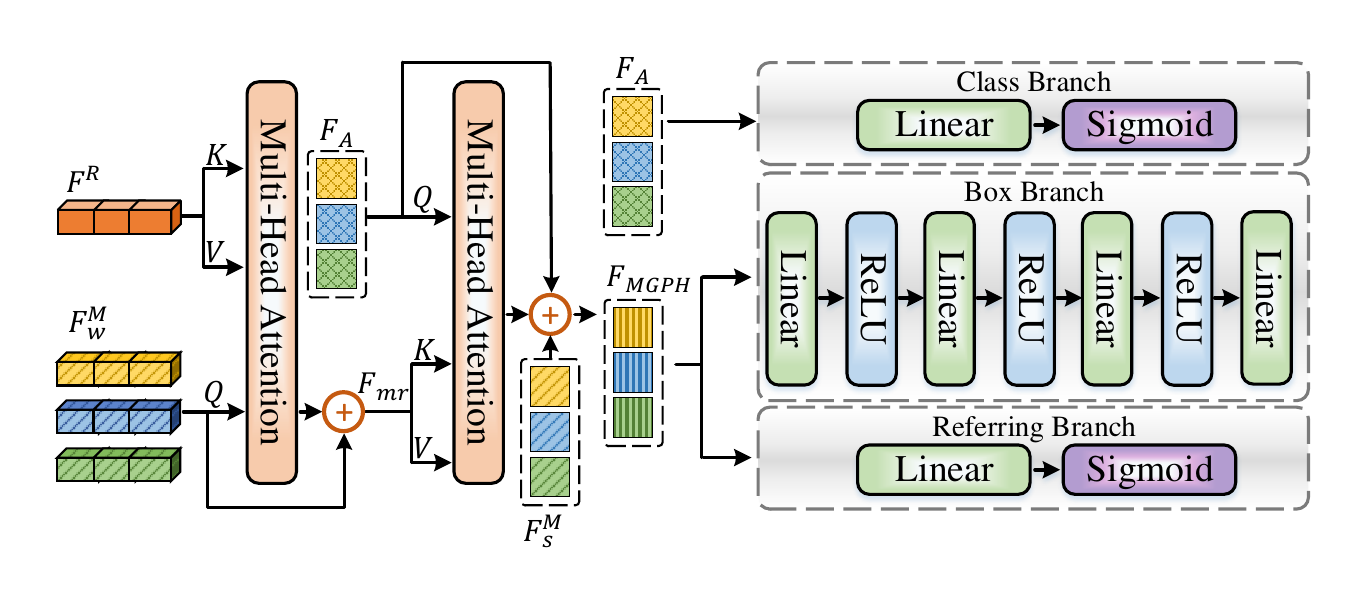}

\end{overpic}
\vspace{-10pt}
\caption{Illustration of MGPH, where motion and reference embeddings are fused to produce motion- and reference-aware features for prediction. 
MGPH consists of three branches: a class branch, a box branch, and a referring branch.}
\label{fig4}
\end{figure}

\subsection{Motion-Guided Prediction Head}
\label{sec:3.4}
To accurately predict object states while leveraging both motion and reference cues, each query first passes through a Temporal Enhancement Module, following TempRMOT \cite{(11)temprmot}, to strengthen its temporal perception 
from historical frames.
The temporally enhanced query is then fed into the MGPH, 
where motion and reference embeddings are fused to produce motion- and reference-aware features for prediction.

As shown in Fig.~\ref{fig4}, in MGPH, the motion- and reference-aware features are fused with the updated track query $\mathbf{F}_{A}$ through multi-head cross-attention operations. 
Specifically, the reference features $\mathbf{F}^{R}$, along with the sentence-level and word-level motion features $\mathbf{F}^{M}_s$ and $\mathbf{F}^{M}_w$, are integrated with the track query as follows.

First, the word-level motion features $\mathbf{F}^{M}_w$ are aligned with the reference features $\mathbf{F}^{R}$ through cross-attention, producing a motion-reference fused representation:
\begin{equation}
\mathbf{F}_{mr} = \operatorname{MHA}(\mathbf{F}^{M}_w, \mathbf{F}^{R}, \mathbf{F}^{R}) + \mathbf{F}^{M}_w,
\label{eq9}
\end{equation}

Next, the track query $\mathbf{F}_{A}$ attends to the fused motion-reference feature $\mathbf{F}_{mr}$, integrating both motion and reference information while retaining the global motion context through a residual connection with the sentence-level motion embedding $\mathbf{F}^{M}_s$. 
This design allows the query to simultaneously leverage local temporal cues, overall motion trends, and linguistic references, improving the accuracy and consistency of trajectory prediction. The final motion- and reference-aware feature is computed as
\begin{equation}
\mathbf{F}_{MGPH} = \operatorname{Linear}( \operatorname{MHA}(\mathbf{F}_{A}, \mathbf{F}_{mr}, \mathbf{F}_{mr}) + \mathbf{F}_{A} + \mathbf{F}^{M}_s ),
\end{equation}
where $\operatorname{Linear}$ denotes a linear layer, and $\mathbf{F}_{MGPH}$ is subsequently used for bounding box regression and reference confidence prediction.

Finally, the MGPH predicts the object class, reference alignment confidence, and bounding box offsets. 
The class branch applies a linear layer with a sigmoid activation to produce a confidence score, indicating whether a query represents a valid object. The referring branch similarly uses a linear layer with sigmoid activation to evaluate the alignment between the query and the language reference. The box branch employs a compact multi-layer perceptron to predict bounding box offsets, which are then added to the prior box estimates to yield refined object positions.
The resulting object states are subsequently stored in the memory bank $\mathcal{M}$ to provide historical motion information for future frames.

\section{Experiments}
\label{sec4}

\begin{table*}[t]
\centering
\footnotesize
\renewcommand{\arraystretch}{1.15}

\caption{
Comparison with state-of-the-art RMOT trackers on the Refer-KITTI test set.
* denotes the results after frame correction.
‡ represents the results obtained by directly evaluating the official open-source code and released weights.
\textbf{E} indicates end-to-end methods.
$\uparrow$ denotes higher is better.
\textbf{Bold} numbers indicate the best results.}

\resizebox{\textwidth}{!}{
\begin{tabular}{lccccccccccccc}
\toprule

\textbf{Method}
& \textbf{Publication}
& \textbf{Detection}
& \textbf{E}
& \textbf{HOTA}$\uparrow$
& \textbf{DetA}$\uparrow$
& \textbf{AssA}$\uparrow$
& \textbf{DetRe}$\uparrow$
& \textbf{DetPr}$\uparrow$
& \textbf{AssRe}$\uparrow$
& \textbf{AssPr}$\uparrow$
& \textbf{MOTA}$\uparrow$
& \textbf{IDF1}$\uparrow$ \\

\midrule

FairMOT \cite{(1)Fairmot}
& IJCV2021
& DLA-34
& \ding{55} & 23.46 & 14.84 & 40.15 & 17.40 & 43.58 & 53.35 & 73.15 & 0.80 & 26.18 \\

DeepSORT \cite{(2)deepsort}
& ICIP2017
& DLA-34
& \ding{55} & 25.59 & 19.76 & 34.31 & 26.38 & 36.93 & 39.55 & 61.05 & -- & -- \\

ByteTrack \cite{(3)bytetrack}
& ECCV2022
& DLA-34
& \ding{55} & 22.49 & 13.17 & 40.62 & 16.13 & 36.61 & 46.09 & 73.39 & -7.52 & 23.72 \\

CSTrack \cite{(4)CSTrack}
& TIP2022
& YOLOv5
& \ding{55} & 27.91 & 20.65 & 39.10 & 33.76 & 32.61 & 43.12 & 71.82 & -- & -- \\

TransTrack \cite{(5)transtrack}
& arXiv2020
& DeformableDETR
& \ding{55} & 32.77 & 23.31 & 45.71 & 32.33 & 42.23 & 49.99 & 79.48 & -- & -- \\

TrackFormer \cite{(6)trackformer}
& CVPR2022
& DeformableDETR
& \ding{55} & 33.26 & 25.44 & 45.87 & 35.21 & 42.19 & 50.26 & 79.63 & -- & -- \\

iKUN \cite{(7)ikun}
& CVPR2024
& DeformableDETR
& \ding{55} & 48.84 & 35.74 & 66.80 & 51.97 & 52.25 & 72.95 & 87.09 & 12.26 & 54.05 \\ 

ReferGPT$_{3D}$ \cite{(31)refergpt}
& CVPR2025
& CasA
& \ding{55} & 49.46 & 39.43 & 62.57 & 50.21 & 58.91 & 73.74 & 72.78 & - & - \\

\midrule

EchoTrack \cite{(8)echotrack}
& TITS2024
& DeformableDETR
& \ding{51} & 39.47 & 31.19 & 51.56 & 42.65 & 48.86 & 56.68 & 81.21 & -- & -- \\

DeepRMOT \cite{(9)deeprmot}
& ICASSP2024
& DeformableDETR
& \ding{51} & 39.55 & 30.12 & 53.23 & 41.91 & 47.47 & 58.47 & 82.16 & -- & -- \\

TransRMOT \cite{(10)transrmot}
& CVPR2023
& DeformableDETR
& \ding{51} & 38.06 & 29.28 & 50.83 & 40.20 & 47.36 & 55.43 & 81.36 & 9.03 & 46.40 \\

TransRMOT\textsuperscript{*} \cite{(10)transrmot}
& CVPR2023
& DeformableDETR
& \ding{51} & 46.56 & 37.97 & 57.33 & 49.69 & \textbf{60.10} & 60.02 & \textbf{89.67} & 17.88 & 53.60 \\

MLS-Track \cite{(25)mlstrack}
& arXiv2024
& DeformableDETR
& \ding{51} & 49.05 & 40.03 & 60.25 & 59.07 & 54.18 & 65.12 & 88.12 & - & - \\

CDRMT \cite{(26)CDRMT}
& \makecell{Information\\Fusion 2025}
& DeformableDETR
& \ding{51} & 49.35 & 40.34 & 60.56 & 54.54 & 59.30 & 64.70 & 89.80 & - & - \\

TenRMOT \cite{(28)tenrmot}
& TCSVT2025
& DeformableDETR
& \ding{51} & 49.77 & 40.79 & 60.89 & 52.65 & 62.81 & 65.38 & 89.28 & - & - \\

TempRMOT\textsuperscript{‡} \cite{(11)temprmot}
& arXiv2024
& DeformableDETR
& \ding{51} & 51.40 & 39.72 & 66.67 & 53.55 & 59.13 & 71.98 & 87.24 & 24.84 & 58.75 \\

DKGTrack \cite{(24)dkgtrack}
& CVPR2025
& DeformableDETR
& \ding{51} & 52.08 & 41.10 & 66.04 & \textbf{57.57} & 58.36 & 71.13 & 87.98 & - & - \\

\midrule

VMRMOT
& Ours
& DeformableDETR
& \ding{51} & \textbf{53.00} & \textbf{41.13} & \textbf{68.41} & 55.63 & 59.74 & \textbf{73.28} & 88.61 & \textbf{26.66} & \textbf{61.06} \\

\bottomrule
\end{tabular}
}

\label{T1}
\end{table*}

\begin{table*}[t]
\centering
\footnotesize
\renewcommand{\arraystretch}{1.15}

\caption{
Comparison with state-of-the-art RMOT trackers on the Refer-KITTI-V2 test set.
‡ represents the results obtained by directly evaluating the official open-source code and released weights.
\textbf{E} indicates end-to-end methods.
$\uparrow$ denotes higher is better.
\textbf{Bold} numbers indicate the best results.
}

\begin{tabular}{lcccccccccc}
\toprule

\textbf{Method}
& \textbf{Publication}
& \textbf{Detection}
& \textbf{E}
& \textbf{HOTA}$\uparrow$
& \textbf{DetA}$\uparrow$
& \textbf{AssA}$\uparrow$
& \textbf{DetRe}$\uparrow$
& \textbf{DetPr}$\uparrow$
& \textbf{AssRe}$\uparrow$
& \textbf{AssPr}$\uparrow$ \\

\midrule

FairMOT \cite{(1)Fairmot}
& IJCV2021
& DLA-34
& \ding{55} & 22.53 & 15.80 & 32.82 & 20.60 & 37.03 & 36.21 & 71.94 \\

ByteTrack \cite{(3)bytetrack}
& ECCV2022
& DLA-34
& \ding{55} & 24.59 & 16.78 & 36.63 & 22.60 & 36.18 & 41.00 & 69.63 \\

iKUN \cite{(7)ikun}
& CVPR2024
& DeformableDETR
& \ding{55} & 10.32 & 2.17 & 49.77 & 2.36 & 19.75 & 58.48 & 68.64 \\

ReferGPT$_{3D}$ \cite{(31)refergpt}
& CVPR2025
& CasA
& \ding{55} & 30.12 & 15.69 & 59.02 & 21.55 & 34.41 & 74.59 & 68.20 \\

\midrule

TransRMOT \cite{(10)transrmot}
& CVPR2023
& DeformableDETR
& \ding{51} & 31.00 & 19.40 & 49.68 & 36.41 & 28.97 & 54.59 & \textbf{82.29} \\

CDRMT \cite{(26)CDRMT}
& \makecell{Information\\Fusion 2025}
& DeformableDETR
& \ding{51} & 31.99 & 20.37 & 50.35 & 26.40 & 46.26 & 53.40 & 85.90 \\

TempRMOT\textsuperscript{‡} \cite{(11)temprmot}
& arXiv2024
& DeformableDETR
& \ding{51} & 34.05 & 22.22 & \textbf{52.50} & 32.58 & \textbf{40.20} & \textbf{58.17} & 81.65 \\

\midrule

VMRMOT
& Ours
& DeformableDETR
& \ding{51} & \textbf{35.21} & \textbf{23.96} & 51.95 & \textbf{37.03} & 39.59 & 57.56 & 81.34 \\

\bottomrule
\end{tabular}

\label{T2}
\end{table*}

\subsection{Experimental Setup}
\textbf{Datasets.}
We conducted the main experiments on Refer-KITTI~\cite{(10)transrmot} and Refer-KITTI-V2~\cite{(11)temprmot} datasets.
Refer-KITTI is the first publicly available dataset for RMOT, derived from the KITTI~\cite{(12)kitti} benchmark with manually annotated referring expressions, which includes 15 training videos with 660 descriptions and 3 test videos with 158 annotations.
Refer-KITTI-V2 is an extension of Refer-KITTI, expanding the dataset with the full KITTI sequences and replacing manual annotations with LLM-generated referring expressions, resulting in 17 training videos with 8,873 annotations and 4 test videos with 897 annotations.

\textbf{Evaluation Metrics.}
We utilize Higher Order Metric \cite{(13)hota} (HOTA, DetA, AssA), IDF1 \cite{(14)idf1}, and CLEAR metrics \cite{(15)mota} (MOTA) as our main evaluation metrics.
HOTA serves as the primary metric, balancing detection and association accuracy. IDF1 and AssA measure association performance, while DetA and MOTA mainly assess detection quality.
Notably, in RMOT, an object is regarded as a true positive (TP) only when it correctly matches the reference description. Therefore, MOTA, which is heavily influenced by false positives (FP) and false negatives (FN), intuitively reflects the model’s ability to align visual objects with references.

\textbf{Implementation Details.}
Our training process is conducted in two stages. First, we follow the training setup of TempRMOT \cite{(11)temprmot} to pre-train the transformer for $60$ epochs. 
Then, the transformer is frozen, and the entire framework, including VMRA, MGPH, and the MLLM components, is jointly fine-tuned. The MLLM module leverages LoRA \cite{(49)lora} for efficient fine-tuning of the text encoder and cross-modal fusion layers.
For optimization in the second stage, we adopt the AdamW optimizer with an initial learning rate of $10^{-5}$, and fine-tune the model for $5$ epochs, where the learning rate is decayed by a factor of $10$ at the $4^{th}$ epoch. The batch size is set to $1$. 
All experiments are conducted on four NVIDIA GeForce RTX 3090 GPUs.
We employ the Qwen2.5-Omni-3B \cite{(16)Qwen2.5-Omni} model as our MLLM, and all input image sequences fed into the model are resized to $336 \times 336$.
The LoRA rank is 16.
The memory bank size $n$ of $\mathcal{M}$ is set to 4 for Refer-KITTI and 5 for Refer-KITTI-V2. 
The feature dimension $D$ is 256.
During inference, objects with a class confidence above $0.6$ and a referring confidence above $0.4$ are regarded as successfully detected objects.

\begin{table}[t]
\renewcommand{\arraystretch}{1.3}
\begin{center}
\caption{ 
Ablation studies on the Refer-KITTI dataset to analyze the contribution of different modules in VMRMOT.
}
\begin{tabular}{c   c   c  | c   c   c}
\toprule

Base  & VMRA  & MGPH  & 
HOTA$\uparrow$ & DetA$\uparrow$ & AssA$\uparrow$ \\

\midrule
\ding{51} &   &   & 51.40 & 39.72 & 66.67\\
\ding{51} & \ding{51}  &   & 52.75 & 41.06 & 67.91\\
\ding{51} &   & \ding{51}  & 52.33 & 40.95 & 67.02\\
\ding{51} & \ding{51}  & \ding{51}  & \textbf{53.00} & \textbf{41.13} & \textbf{68.41} \\

\bottomrule
\end{tabular}

\label{T3}
\vspace{-10pt}
\end{center}
\end{table}

\begin{table}[t]
\renewcommand{\arraystretch}{1.3}
\begin{center}
\caption{
Comparison of different motion feature extractors on the Refer-KITTI dataset.
}
\begin{tabular}{l | c c c}
\toprule

Motion Feature Extractor &
HOTA$\uparrow$ & DetA$\uparrow$ & AssA$\uparrow$ \\

\midrule

No motion feature & 51.40 & 39.72 & 66.67 \\

RoBERTa & 52.46 & 40.74 & 67.70 \\

MiniCPM-V-2 & 52.67 & 41.05 & 67.73 \\

Qwen2.5-VL-3B & 52.73 & 41.01 & 67.97 \\

Qwen2.5-Omni-3B & \textbf{53.00} & \textbf{41.13} & \textbf{68.41} \\

\bottomrule
\end{tabular}

\label{T4}
\vspace{-10pt}
\end{center}
\end{table}

\begin{table}[t]
\renewcommand{\arraystretch}{1.3}
\begin{center}
\caption{
Comparison of different motion representations on the Refer-KITTI dataset. 
``Numeric'' encodes motion as numerical vectors via a Transformer, while ``Descriptive'' converts motion into natural-language descriptions processed by the MLLM.  
The best results are shown in bold.
}
\begin{tabular}{c |  c  c   c}

\toprule
Motion Representation &

HOTA$\uparrow$ & DetA$\uparrow$ & AssA$\uparrow$ \\

\midrule
Numeric  & 52.34 & 40.72 & 67.42 \\
Descriptive & \textbf{53.00} & \textbf{41.13} & \textbf{68.41} \\
\bottomrule
\end{tabular}

\label{ST2}
\vspace{-10pt}
\end{center}
\end{table}

\subsection{Benchmark Evaluation}

\textbf{Refer-KITTI.}
We report VMRMOT's performance on Refer-KITTI in Tab.~\ref{T1}.
Following TransRMOT, the reported FairMOT, DeepSORT, ByteTrack, CStrack, TransTrack, and TrackFormer results correspond to their RMOT adaptations equipped with cross-modal fusion modules rather than the original MOT implementations.
It can be seen that VMRMOT achieves the best overall performance with $53.00\%$ HOTA, $41.13\%$ DetA, $68.41\%$ AssA, $26.66\%$ MOTA, and $61.06\%$ IDF1.
Compared with the end-to-end transformer-based RMOT baselines TransRMOT \cite{(10)transrmot} and TempRMOT \cite{(11)temprmot}, VMRMOT consistently improves tracking performance. Specifically, VMRMOT surpasses TransRMOT by $6.44\%$ in HOTA, $3.16\%$ in DetA, and $11.08\%$ in AssA, while outperforming TempRMOT by $1.60\%$, $1.41\%$, and $1.74\%$ in these metrics, respectively. Furthermore, compared with the recent state-of-the-art RMOT method DKGTrack \cite{(24)dkgtrack}, VMRMOT still achieves the best overall performance, improving HOTA, DetA, and AssA by $0.92\%$, $0.03\%$, and $2.37\%$, respectively. These results demonstrate that the proposed motion-aware vision-reference alignment framework effectively enhances both detection and association performance in RMOT.
In addition, it is worth noting that VMRMOT also achieves consistent improvements in the MOTA metric.
Specifically, VMRMOT surpasses TransRMOT by $8.78\%$ and TempRMOT by $1.82\%$ in MOTA. Since MOTA is directly affected by FP and FN, this improvement provides more intuitive evidence that introducing the motion modality helps VMRMOT achieve better alignment between the reference and vision modalities.

\textbf{Refer-KITTI-V2.}
To further validate the effectiveness of VMRMOT, we evaluate it on the Refer-KITTI-V2 dataset, which includes more linguistically diverse yet static references.
As shown in Tab.~\ref{T2}, VMRMOT achieves the best performance with $35.21\%$ HOTA and $23.96\%$ DetA, as well as the second-best $51.95\%$ AssA.
Compared with TransRMOT \cite{(10)transrmot} and TempRMOT \cite{(11)temprmot}, VMRMOT yields gains of $4.21\%$ and $1.16\%$ in HOTA, respectively.
These results demonstrate the effectiveness of introducing the motion modality to enhance vision–reference alignment, enabling VMRMOT to maintain strong multi-modal tracking performance even under more complex references.

\begin{table}[t]
\renewcommand{\arraystretch}{1.3}
\begin{center}
\caption{ 
Comparison of different architectures of VMRA on the Refer-KITTI dataset.
}

\begin{tabular}{c   c   c  | c   c   c}
\toprule

$\mathbf{F}^{M}_w$  & $\mathbf{F}^{M}_s$  & $\mathbf{F}^{R}$  & 
HOTA$\uparrow$ & DetA$\uparrow$ & AssA$\uparrow$ \\

\midrule
\ding{51} &  &  &   52.37 & 40.46  & 67.94 \\
\ding{51} & \ding{51}  &   & 52.54 & 41.12 & 67.27\\
 &   & \ding{51}  & 52.12  & 40.67  & 66.94 \\
\ding{51} & \ding{51}  & \ding{51}  & \textbf{53.00} & \textbf{41.13} & \textbf{68.41} \\
\bottomrule
\end{tabular}

\label{T5}
\vspace{-10pt}
\end{center}
\end{table}

\begin{table}[t]
\renewcommand{\arraystretch}{1.3}
\begin{center}

\caption{ 
Comparison of different architectures of MGPH on the Refer-KITTI dataset.
}
\begin{tabular}{c   c  | c   c   c}
\toprule

$\mathbf{F}^{M}_w$  & $\mathbf{F}^{R}$  & 
HOTA$\uparrow$ & DetA$\uparrow$ & AssA$\uparrow$ \\

\midrule
\ding{51} &   & 52.85 & \textbf{41.25} & 67.85\\
 & \ding{51} & 51.97 & 40.59 & 66.67\\
\ding{51} & \ding{51} & \textbf{53.00} & 41.13 & \textbf{68.41} \\

\bottomrule
\end{tabular}

\label{T6}
\vspace{-10pt}
\end{center}
\end{table}

\begin{table*}[t]
\centering
\renewcommand{\arraystretch}{1.3}
\caption{
Robustness analysis under noisy historical trajectories on the Refer-KITTI dataset.
To simulate localization drift, Gaussian noise is injected into the center coordinates of historical trajectory observations stored in the Memory Bank:
$(c_x,c_y)\leftarrow(c_x,c_y)+\mathcal{N}(0,\sigma)$.
The percentage in the Setting column indicates the ratio of historical trajectory observations that are randomly corrupted by the injected noise.
Description Accuracy measures the consistency between motion descriptions generated from noisy trajectories and those generated from clean trajectories.
For each object, we compare four motion attributes (position descriptor, motion direction descriptor, distance trend descriptor, and speed trend descriptor), and compute the percentage of matched attributes.
}
\begin{tabular}{lccccc}
\toprule
\textbf{Noise Type}
& \textbf{Setting}
& \textbf{Description Accuracy (\%)}
& \textbf{HOTA}$\uparrow$
& \textbf{DetA}$\uparrow$
& \textbf{AssA}$\uparrow$ \\
\midrule

None (Clean)
& -- & 100.0 & 53.00 & 41.13 & 68.41 \\

\midrule

Drift
& $\sigma=0.05$, 25\% & 83.0 & 52.90 & 41.14 & 68.16 \\

Drift
& $\sigma=0.10$, 25\% & 82.3 & 52.92 & 41.17 & 68.18 \\

Drift
& $\sigma=0.20$, 25\% & 82.1 & 52.88 & 41.04 & 68.28 \\

Drift
& $\sigma=0.05$, 50\% & 72.0 & 52.94 & 41.09 & 68.33 \\

\bottomrule
\end{tabular}
\label{MR2T1}
\end{table*}

\begin{table}[t]
\renewcommand{\arraystretch}{1.3}
\caption{
Inference efficiency comparison on the Refer-KITTI test set evaluated on a single RTX 3090 GPU.
VMRMOT-P$K$ denotes invoking the MLLM once every $K$ frames.
}
\begin{center}
\begin{tabular}{l   |  c   c   c}
\toprule

Method & GFLOPs  & HOTA & FPS\\

\midrule
TransRMOT \cite{(10)transrmot} & \textbf{136.35} & 46.56 & \textbf{11.68} \\
TempRMOT \cite{(11)temprmot} & 139.70 & 51.40 & 10.26 \\
\hline
VMRMOT & 242.55 & \textbf{53.00} & 1.65 \\
VMRMOT-P2 & 190.99 & 52.58 & 2.52 \\
VMRMOT-P5 & 160.27 & 52.56 & 4.85 \\
VMRMOT-P10 & 149.99 & 52.48 & 8.87 \\

\bottomrule
\end{tabular}

\label{RT1}
\vspace{-10pt}
\end{center}
\end{table}

\begin{table}[t]
\renewcommand{\arraystretch}{1.3}
\caption{
Lightweight deployment analysis on Refer-KITTI is evaluated on a single RTX 3090 GPU.
The lightweight configuration reduces the transformer size by setting
$\texttt{enc\_layers}$ (encoder layers) from 6 to 3,
$\texttt{dec\_layers}$ (decoder layers) from 6 to 3,
$\texttt{dim\_feedforward}$ (feed-forward dimension) from 1024 to 512,
and $\texttt{num\_feature\_levels}$ (multi-scale feature levels) from 4 to 3.
VMRMOT-P$K$ denotes invoking the motion encoder once every $K$ frames.
}
\begin{center}
\begin{tabular}{l | c c}
\toprule
Method & HOTA & FPS\\
\midrule

TempRMOT-Small \cite{(11)temprmot} & 33.23 & \textbf{19.21} \\

\hline

VMRMOT-Small & \textbf{35.45} & 2.14 \\
VMRMOT-Small-P2 & 35.33 & 4.91 \\
VMRMOT-Small-P5 & 34.26 & 10.71 \\
VMRMOT-Small-P10 & 34.35 & 16.49 \\

\bottomrule
\end{tabular}
\label{MR1T4}
\vspace{-10pt}
\end{center}
\end{table}

\begin{figure*}[tb]
\centering
\begin{overpic}
[width=1\linewidth,]{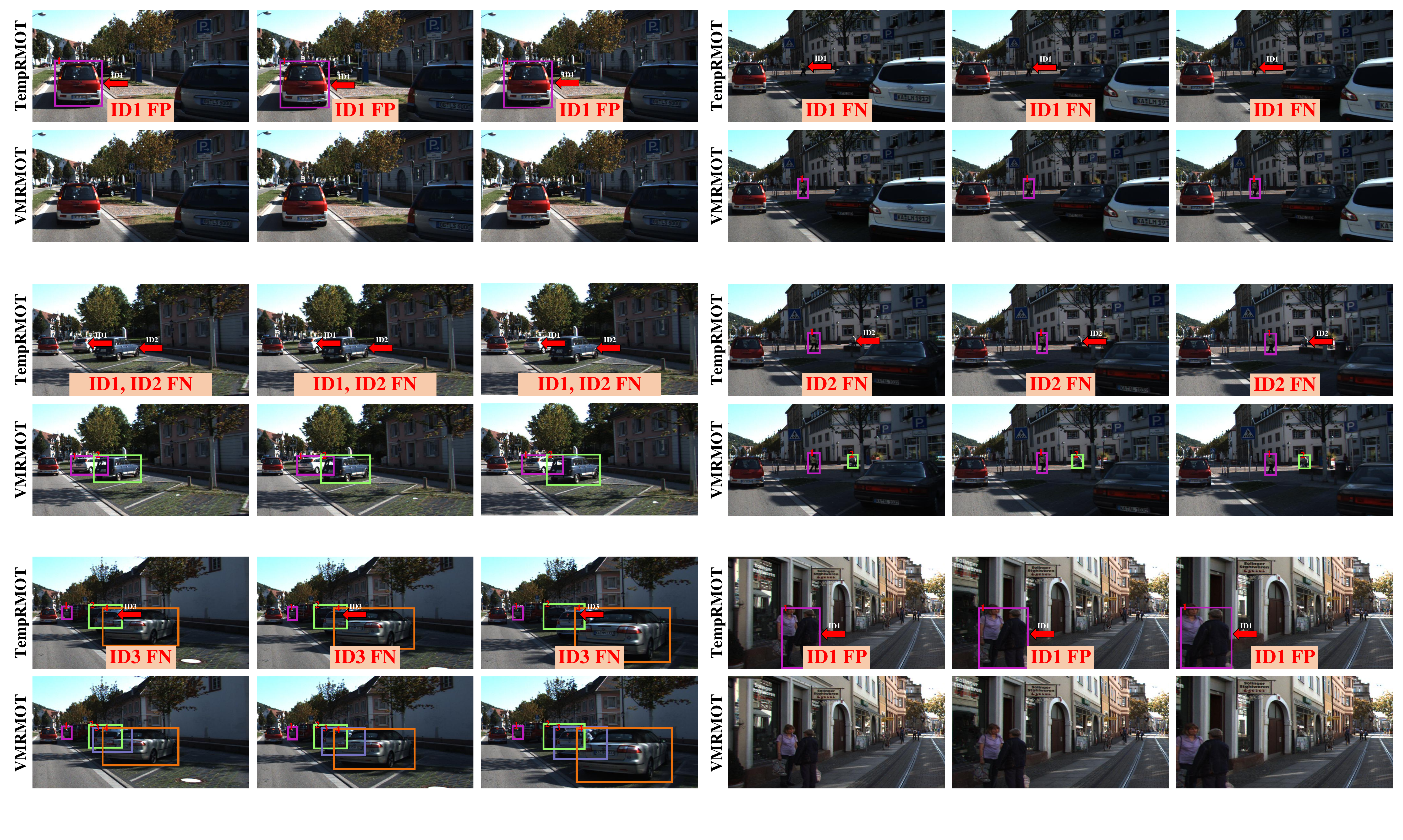}
\put(13, 40.5){(a) Reference: "Moving Cars"}
\put(62, 40.5){(b) Reference: "Moving Pedestrain"}
\put(11, 21){(c) Reference: "Parking Vehicles"}
\put(58, 21){(d) Reference: "People Who Are Walking"}
\put(7, 1){(e) Reference: "Right Cars Which Are Parking"}
\put(55, 1){(f) Reference: "Left Pedestrian Who Are walking"}

\end{overpic}
\caption{Qualitative comparison between TempRMOT and VMRMOT on the Refer-KITTI dataset.
The red arrow indicates the noteworthy objects.
Boxes of the same color represent the same ID.
Best viewed in color and zoom-in.
}
\vspace{-10pt}
\label{SF1}
\end{figure*}

\begin{figure*}[tb]
\centering
\begin{overpic}
[width=1\linewidth,]{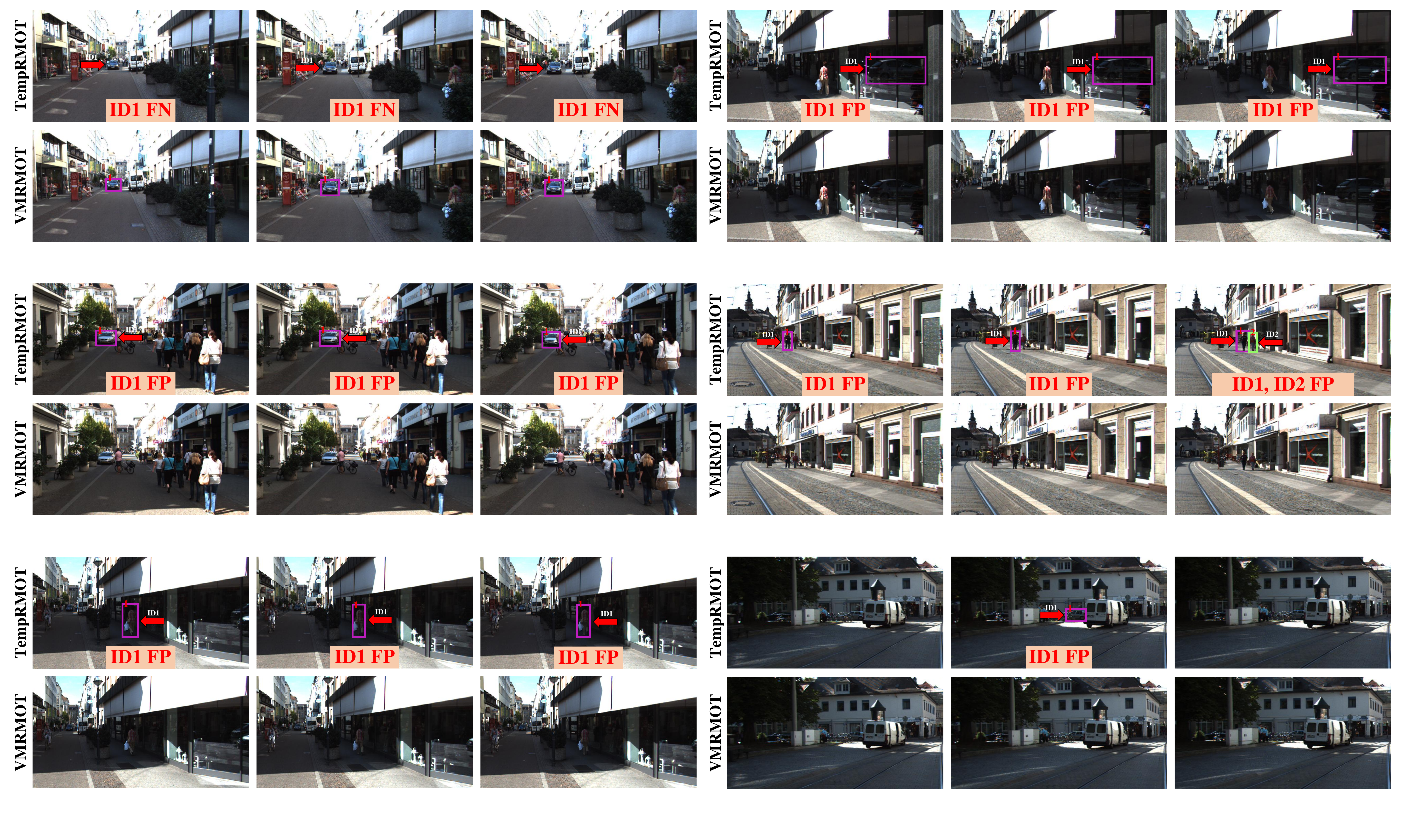}
\put(10, 40.5){(a) Reference: "Autos Parked Are Black"}
\put(53.5, 40.5){(b) Reference: "Autos That Are Black Are In Motion"}
\put(13, 21){(c) Reference: "Cars In Motion"}
\put(52.5, 21){(d) Reference: "Individuals Are Standing On The Right"}
\put(21, 2){(e) Reference:}
\put(1.5, 0){"Someone In A Coat And Jeans Walking On The Sidewalk"}
\put(71, 2){(f) Reference:}
\put(56, 0){ "Parked Cars That Are Silver Were Left-behind"}

\end{overpic}
\caption{Qualitative comparison between TempRMOT and VMRMOT on the Refer-KITTI-V2 dataset.
The red arrow indicates the noteworthy objects.
Boxes of the same color represent the same ID.
Best viewed in color and zoom-in.
}
\label{SF2}
\end{figure*}

\subsection{Ablation Studies}
We conduct ablation studies on the Refer-KITTI dataset.
The ablation studies focus on analyzing the contribution of different modules in VMRMOT, different motion feature extractors, the impact of motion representation form, different architectures of VMRA, and different architectures of MGPH.

\textbf{Contribution of Different Modules.}
To analyze the contribution of the proposed VMRA and MGPH modules, we conduct an ablation study, as shown in Tab.~\ref{T3}.
In the table, “Base” refers to the baseline model without VMRA and MGPH, which corresponds to TempRMOT \cite{(11)temprmot}.
From the table, incorporating VMRA/MGPH achieves $52.75\%$/$52.33\%$ HOTA, $41.06\%$/$40.95\%$ DetA, and $67.91\%$/$67.02\%$ AssA, improving the baseline by $1.35\%$/$0.93\%$, $1.34\%$/$1.23\%$, and $1.24\%$/$0.35\%$, respectively.
The results demonstrate that incorporating the motion modality enhances vision–reference alignment, and utilizing motion cues in the prediction head further strengthens multi-modal tracking.
Moreover, combining both VMRA and MGPH achieves the best performance, reaching $53.00\%$ HOTA, $41.13\%$ DetA, and $68.41\%$ AssA, demonstrating that both modules contribute significantly to the overall multi-modal tracking performance of VMRMOT.

\textbf{Different Methods to Extract Motion Features.}
To investigate the impact of different motion feature extraction methods, we conduct experiments as shown in Tab.~\ref{T4}.
In the table, “No motion feature” denotes the baseline model without incorporating the motion modality.
“RoBERTa” represents extracting motion features directly from motion descriptions using a pretrained RoBERTa model.
“Qwen2.5-Omni-3B” refers to extracting motion features from both motion descriptions and visual sequences using the Qwen2.5-Omni-3B model, which is used in VMRMOT.
As shown in the results, the “RoBERTa” method achieves $52.46\%$ HOTA, surpassing the baseline by $1.06\%$.
This demonstrates that introducing a motion modality derived from motion descriptions indeed enhances cross-modal alignment and tracking performance, as the extracted motion features encode dynamic behaviors of objects.
Furthermore, the “Qwen2.5-Omni-3B” method produces motion features that better model the alignment between motion descriptions and visual dynamics through an MLLM, leading to the best overall tracking performance.

\textbf{Different Motion Feature Extractors.}
To investigate the impact of different motion feature extractors, we conduct experiments as shown in Tab.~\ref{T4}.
In the table, "No motion feature" denotes the baseline model without incorporating the motion modality.
"RoBERTa" represents extracting motion features directly from motion descriptions using a pretrained RoBERTa model.
``MiniCPM-V-2'', ``Qwen2.5-VL-3B'', and ``Qwen2.5-Omni-3B'' represent extracting motion features using different multimodal large language models.
As shown in the results, the ``RoBERTa'' method achieves $52.46\%$ HOTA, surpassing the baseline by $1.06\%$.
This demonstrates that introducing a motion modality derived from motion descriptions indeed enhances cross-modal alignment and tracking performance, as the extracted motion features encode dynamic behaviors of objects.
Furthermore, both MiniCPM-V-2 and Qwen2.5-VL-3B consistently outperform RoBERTa, achieving $52.67\%$ and $52.73\%$ HOTA, respectively. These results indicate that the proposed framework is compatible with different motion feature extractors and does not rely on a specific MLLM architecture. Different MLLMs can effectively encode the proposed motion-aware descriptions and consistently improve tracking performance.
Among all evaluated motion feature extractors, Qwen2.5-Omni-3B achieves the best overall performance, obtaining $53.00\%$ HOTA, $41.13\%$ DetA, and $68.41\%$ AssA. This suggests that stronger multimodal representation capabilities enable more effective modeling of the correspondence between motion descriptions and visual dynamics, leading to more discriminative motion features and better vision--motion--reference alignment.

\textbf{Impact of Motion Representation Form.}
We evaluate how different formulations of motion information affect tracking performance. 
As shown in Tab.~\ref{ST2}, representing motion as numeric vectors and encoding them with a Transformer yields $52.34\%$ HOTA.
This approach provides only moderate improvements because raw numerical inputs lack sufficient semantic expressiveness.
In contrast, converting motion into natural-language descriptions and encoding them with the MLLM leads to a clear performance gain, achieving $53.00\%$ HOTA.
Compared with directly using low-level trajectory representations, the proposed motion-aware descriptions achieve better performance. This result indicates that representing object dynamics as an explicit motion modality is more effective than directly operating on geometric trajectories. The improvement demonstrates that the proposed framework benefits not only from trajectory information itself, but also from reformulating object dynamics into motion-aware representations that facilitate vision--reference alignment.
This improvement suggests that descriptive motion language provides richer temporal and semantic cues, allowing the MLLM to better align motion information with visual evidence and thus enhance multi-modal tracking.

\textbf{Different Architectures of VMRA.}
To evaluate the effectiveness of the VMRA architecture, we conduct experiments as shown in Tab.~\ref{T5}.
In the table, $\mathbf{F}^{M}_w$, $\mathbf{F}^{M}_s$, and $\mathbf{F}^{R}$ denote whether word-level motion features, sentence-level motion features, and reference features are incorporated into the VMRA design, respectively.
As shown, integrating word-level motion features yields $52.37\%$ HOTA, while adding sentence-level motion features further improves performance to $52.54\%$ HOTA. 
When both motion and reference features are jointly fused, the model achieves the best result of $53.00\%$ HOTA, demonstrating the effectiveness of the hierarchical fusion strategy in aligning multi-modal features.
This hierarchical structure allows VMRA to capture both fine-grained motion cues and global temporal semantics, thereby enhancing multi-modal alignment and overall tracking performance.
In contrast, using only reference features without motion features results in a lower $52.12\%$ HOTA, indicating that static reference descriptions alone are insufficient for modeling temporal dynamics.
The results demonstrate that incorporating motion features at multiple levels is crucial for facilitating stronger vision–reference alignment and more accurate tracking performance.

\textbf{Different Architectures of MGPH.}
To validate the effectiveness of the MGPH design, we conduct experiments as shown in Tab.~\ref{T6}.
In the table, $\mathbf{F}^{M}_w$ and $\mathbf{F}^{R}$ indicate whether motion features and reference features are incorporated in the MGPH design, respectively.
As shown, incorporating motion features yields $52.85\%$ HOTA, while further integrating reference features achieves the best result of $53.00\%$ HOTA.
This demonstrates that motion features effectively enhance the performance of the prediction head by providing dynamic behavioral cues of objects, and additionally, fusing reference features introduces complementary reference information that further improves prediction quality.
However, when only reference features are used without motion features, the performance drops to $51.97\%$ HOTA, indicating that reference features mainly contain static descriptions and lack dynamic cues.
Therefore, the motion modality plays a more crucial role in guiding the prediction head by encoding the dynamic behaviors of objects, which are essential for accurate multi-modal tracking.

\subsection{Robustness to Noisy Historical Trajectories}
Since the proposed motion modality is generated from historical trajectories stored in the Memory Bank, we further evaluate the robustness of VMRMOT under noisy trajectory conditions.

Specifically, we randomly select 25\% or 50\% of the historical trajectory observations stored in the Memory Bank and inject Gaussian noise into their center coordinates:
\begin{equation}
(c_x,c_y)\leftarrow(c_x,c_y)+\mathcal{N}(0,\sigma),
\end{equation}
where $\sigma$ controls the noise magnitude. This setting simulates localization drift caused by detection errors, tracking inaccuracies, and object occlusions.

To quantify the influence of trajectory noise on motion description generation, we introduce a Description Accuracy metric. For each object, we compare four motion attributes, including the position descriptor, motion direction descriptor, distance trend descriptor, and speed trend descriptor, between the noisy trajectory and the corresponding clean trajectory. Description Accuracy is computed as
\begin{equation}
\mathrm{DescAcc}
=
\frac{1}{4N}
\sum_{i=1}^{N}
\sum_{j=1}^{4}
\mathbf{1}
\left[
d_{i,j}^{\mathrm{noisy}}
=
d_{i,j}^{\mathrm{clean}}
\right]
\times 100\% ,
\label{eq:desc_acc}
\end{equation}
where $N$ denotes the number of objects, $d_{i,j}^{\mathrm{noisy}}$ and $d_{i,j}^{\mathrm{clean}}$ denote the $j$-th motion attribute of the $i$-th object generated from noisy and clean trajectories, respectively, and $\mathbf{1}[\cdot]$ is the indicator function.

The results are reported in Tab.~\ref{MR2T1}. It can be observed that trajectory noise significantly affects the generated motion descriptions. For example, under the $\sigma=0.05$, 50\% setting, the Description Accuracy decreases from $100.0\%$ to $72.0\%$. However, the tracking performance remains remarkably stable. Specifically, HOTA only changes from $53.00\%$ to $52.94\%$, while DetA and AssA decrease by only $0.04\%$ and $0.08\%$, respectively.

These results indicate that VMRMOT does not rely on perfectly accurate motion descriptions. Although trajectory noise inevitably introduces errors into the generated motion modality, the resulting impact on tracking performance remains limited. We attribute this robustness to the complementary role of the motion modality, which assists vision--reference alignment rather than serving as the sole source of tracking decisions.

Overall, the results demonstrate that the proposed framework remains highly robust under noisy trajectory conditions. Even when the Description Accuracy drops by $28\%$, the degradation in tracking performance is negligible, indicating that the proposed motion modality can effectively tolerate imperfect historical trajectories in practical scenarios.

\subsection{Inference Efficiency} 
Since incorporating an MLLM introduces additional computational overhead, we conduct inference efficiency analysis in Tab.~\ref{RT1}. In our design, the MLLM is not invoked at every frame. Instead, motion descriptions are generated at the tracklet level and can be computed periodically and cached.

As shown in the table, invoking the MLLM every $K$ frames (e.g., VMRMOT-P2 / P5 / P10) provides a flexible accuracy–efficiency trade-off. Increasing $K$ substantially improves FPS while retaining most of the HOTA gains. Notably, VMRMOT-P10 achieves comparable GFLOPs and FPS to TempRMOT, while delivering superior HOTA. 
These results demonstrate that the proposed motion modality can be integrated with a controllable computational cost.

\subsection{Lightweight Deployment Analysis}
Since RMOT is often intended for real-world surveillance and robotics applications, practical deployment requires both strong tracking performance and efficient inference. Therefore, we design a lightweight version of VMRMOT for deployment-oriented evaluation. Specifically, the lightweight configuration reduces the transformer depth and width by decreasing the number of encoder layers and decoder layers from 6 to 3, the feed-forward dimension from 1024 to 512, and the number of multi-scale feature levels from 4 to 3. The results are shown in Tab.~\ref{MR1T4}.

As shown in the table, the lightweight VMRMOT still consistently outperforms the corresponding TempRMOT baseline. Specifically, VMRMOT-Small achieves 35.45\% HOTA, surpassing TempRMOT-Small by 2.22\%. Although invoking the motion encoder at every frame introduces additional computational overhead, periodically updating motion features provides a favorable accuracy-efficiency trade-off. For example, VMRMOT-Small-P10 achieves 34.35\% HOTA while reaching 16.49 FPS, which is close to the 19.21 FPS of TempRMOT-Small and still maintains a 1.12\% HOTA advantage. These results demonstrate that the proposed motion modality consistently contributes to tracking performance even under lightweight configurations. Furthermore, the framework provides flexible deployment options, enabling different model settings to be adopted according to application-specific requirements on accuracy and efficiency.

\subsection{Visualization}

\textbf{Qualitative Comparison on Refer-KITTI.}
Fig.~\ref{SF1} illustrates additional qualitative comparisons between TempRMOT~\cite{(11)temprmot} and our VMRMOT on the Refer-KITTI dataset. 
As shown, TempRMOT frequently suffers from a large number of FPs and FNs due to the temporal discrepancy between the vision modality and the static references. 
For example, the issue happens on the tracking results by TempRMOT at: (a) ID1 FP; (b) ID1 FN; (c) ID1 FN and ID2 FN; (d) ID2 FN; (e) ID3 FN; (f) ID1 FP.
In contrast, VMRMOT achieves accurate tracking throughout the entire sequence. By introducing the motion modality, VMRMOT effectively bridges the temporal discrepancy between the vision modality and references, enabling robust multi-modal tracking.

\textbf{Qualitative Comparison on Refer-KITTI-V2.}
Fig.~\ref{SF2} illustrates the qualitative comparison between TempRMOT \cite{(11)temprmot} and VMRMOT on the Refer-KITTI-V2 dataset.
Due to the increased linguistic diversity and complexity of the static references in Refer-KITTI-V2, TempRMOT exhibits numerous FP and FN cases, further amplifying the temporal discrepancy between the vision modality and the static references.
For instance, TempRMOT produces erroneous tracking results in: (a) ID1 FN; (b) ID1 FP; (c) ID1 FP; (d) ID1 FP and ID2 FP; (e) ID1 FP; (f) ID1 FP.
In contrast, VMRMOT yields more accurate and stable tracking results throughout the sequence.
By leveraging the motion modality, VMRMOT effectively mitigates the temporal discrepancy and achieves more reliable vision–reference alignment, even under more diverse and challenging references.

\section{Conclusion}
\label{sec5}
In this paper, we present VMRMOT, a novel vision–motion–reference alignment framework for RMOT.
It introduces the motion modality to strengthen alignment between vision and reference modalities. 
Through the proposed VMRA and MGPH, VMRMOT effectively integrates motion cues with visual–reference information, achieving more robust and accurate object tracking.
Extensive experiments on Refer-KITTI and Refer-KITTI-V2 demonstrate that VMRMOT consistently outperforms previous trackers, validating the effectiveness of introducing the motion modality for Referring Multi-Object Tracking.

\ifCLASSOPTIONcaptionsoff
  \newpage
\fi

\bibliographystyle{IEEEtran}
\bibliography{ref.bib}

\end{document}